\documentclass[10pt,letterpaper]{article}
\usepackage{hyperref}
\usepackage{cogsci}
\usepackage{pslatex}
\usepackage{multicol}
\usepackage{apacite}
\usepackage{times}
\usepackage{latexsym}
\usepackage{url}
\usepackage{graphicx}
\usepackage[export]{adjustbox}
\usepackage[toc,page]{appendix}
\usepackage[utf8]{inputenc}
\usepackage[T1]{fontenc}
\usepackage[affil-it]{authblk}

\title{ADAM: A Sandbox for Implementing Language Learning}
\date{}
\author[1]{Ryan Gabbard\thanks{ryan.gabbard@gmail.com. All work done while at ISI, prior to joining Amazon.}}
\author[1]{Deniz Beser\thanks{beser@isi.edu}}
\author[1]{Jacob Lichtefeld\thanks{jacobl@isi.edu}}
\author[1]{Joe Cecil\thanks{cecil@isi.edu}}
\author[2]{Mitch Marcus}
\author[2,3]{Sarah Payne}
\author[2,3]{Charles Yang}
\author[1]{Marjorie Freedman}
\affil[1]{Information Sciences Institute, University of Southern California}
\affil[2]{Department of Computer and Information Science, University of Pennsylvania}
\affil[3]{Department of Linguistics, University of Pennsylvania}

\begin{document}

\maketitle

\begin{abstract}

We present ADAM, a software system for designing and running child language learning experiments in Python. The system uses a virtual world to simulate a grounded language acquisition process in which the language learner utilizes cognitively plausible learning algorithms to form perceptual and linguistic representations of the observed world. The modular nature of ADAM makes it easy to design and test different language learning curricula as well as learning algorithms. In this report, we describe the architecture of the ADAM system in detail, and illustrate its components with examples. We provide our code.\footnote{Our code is available at 
\href{https://github.com/isi-vista/adam}{https://github.com/isi-vista/adam}}

\end{abstract}

\section{Introduction}

ADAM is a software ``sandbox'' for experiments in child language learning.  It enables an experimenter with a modest degree of Python programming experience to quickly and compactly specify a \textbf{curriculum} consisting of a sequence of situations in a virtual world. It then projects these situations into customizable \textbf{linguistic representations} and non-linguistic \textbf{perceptual representations} which can be consumed by learning algorithms.  It also provides a framework for evaluating learner performance both during and after the learning phase. This approach provides several advantages:
\begin{itemize}
    \item Learning from language together with concrete situations captures the fact that humans learn language not from language alone but rather from language grounded in particular situations. 
    \item Using a virtual world for our grounding makes it easy for a researcher to design custom curricula for their experiments (see section \ref{sec:situation-templates}).  
    \item Using a virtual world for grounding provides the learner with a realistic model of a pre-linguistic infant's perception of the world (see section \ref{sec:perceptual-representation}) as grounding, rather than, for example, raw pixels. 
    \item Generating language from situations enables performing experiments across multiple languages by only swapping out one system module (the language generator; see section \ref{sec:language-generator}).
    \item Generating the curriculum and perceptions in Python code makes it easy to perform variations on experiments, including adding irrelevant situations, adding noise to the perceptual representations, and altering word distributions.
\end{itemize}

In the main body of this report, we first review relevant background in the language acquisition literature. Then, we describe the ADAM system architecture at a high level and then provide some examples of what it learns from a simple sample curriculum.  We then describe related work and directions for future work.  Finally, a series of appendices provide detailed descriptions of system components.

\begin{figure*}[tbp]
    \centering
    \includegraphics[scale=0.6]{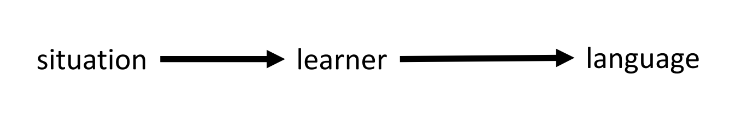}
    \caption{A High-level architecture diagram, which happens in a loop.}
    \label{fig:simplest-architecture}
\end{figure*}

\section{Background}
\label{sec:background}
The current literature consists of several proposals for children's word learning process which fall into two broad groups. `Global' approaches \cite<e.g,>{fazly2010probabilistic, Siskind1996, yu2007, goodman2008, Smith2011}  resolve uncertainty by aggregating situational data over time to identify the best supported  word-referent associations. By contrast, `local' approaches \cite<e.g,>{Medina2011, Spiegel2011, trueswell2013propose, Aravind2018} attempt to resolve uncertainty immediately: in the simplest form, the learner hypothesizes a word meaning, maintains it when the word is heard if it is confirmed (e.g. is present in the observed situation) but replaces it with a new meaning if disconfirmed. Additionally, there are models that combine aspects of both approaches \cite{stevens2017pursuit}, where only actively hypothesized meanings are maintained (as in the local approach), whose associations with the word change probabilistically in response to the cumulative effect of learning instances (as in the global approach).

Most previous computational research has focused on the modeling of behavioral results from word learning experiments or the quantitative assessment of the models' effectiveness on (small) annotated child-directed input corpora. Larger-scale simulation studies   \cite<e.g,>{fazly2010probabilistic, Vogt2012, Blythe2016}, by contrast,  are evaluated on arbitrary word-reference mappings that bear little resemblance to the linguistic, cognitive, and perceptual information available to the child learner. 

Moreover, the words used in previous studies are most often basic-level terms (\emph{dog}), sometimes  reflecting different levels of carefully constructed conceptual taxonomy (`husky', `dog', `animal') \cite<e.g,>{Xu2007, Spencer2011}.  The meanings of the words are almost always atomic.  In the real world, however, word meanings can only be learned by making subtle and incremental refinements. For instance, the word \emph{chair} is used for referents with highly variable attributes (size, color, number of legs, etc.).

ADAM aims to provide a computational platform that enables systematic studies of how the structure of the learning data, the perceptual complexity of the learning environment, and the cognitive resources available to the learner collectively impact the effectiveness of the language acquisition models. In the following sections, we describe the ADAM system in detail.

\section{High-level System Architecture}
\label{sec:system-architecture}

At the highest level, the ADAM system is a loop where a \textbf{language learner} repeatedly perceives \textbf{situations} (states of the world or transitions between such states) paired with linguistic utterances (Figure \ref{fig:simplest-architecture}). After each observation, the learner updates its internal state. The sequence of situations observed by the learner is called a \textbf{curriculum}. At any time, the learner can perceive a new situation and be asked to generate a relevant and appropriate utterance.

\begin{figure*}[tbp]
    \centering
    \includegraphics[scale=0.6]{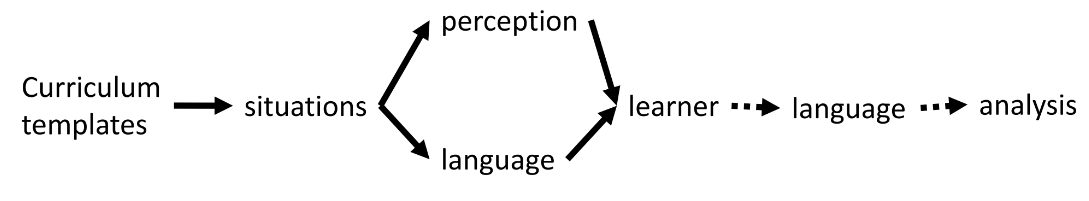}
    \caption{A Mid-level architecture diagram, with additional detail added to the processing loop.}
    \label{fig:mid-architecture}
\end{figure*}

Figure \ref{fig:mid-architecture} displays the full architecture, which has three notable additions. First, the learner has no direct access to the situations in the curriculum.  Instead, it receives only a pairing of a \textbf{linguistic representation} (section \ref{sec:linguistic-representation}) and a \textbf{perceptual representation} (section \ref{sec:perceptual-representation}) which are both generated from a \textbf{situation representation} (section \ref{sec:situation-representation-simple}) by the \textbf{language generator} (section \ref{sec:language-generator}) and \textbf{perception generator} (section \ref{sec:perception-generator}), respectively. In the current implementation, these representations are meant to model an 18-month-old's perception of the world around them. Second, we provide \textbf{situation templates}, which are a compact way for an experimenter to specify a large curriculum (section \ref{sec:situation-templates}).  Third, an \textbf{evaluator} allows for tracking learning progress and evaluating experiments (section \ref{sec:evaluator}). 

Below we describe the components of the system in more detail.

\subsection{Situations and Curricula}
\label{sec:situation-representation-simple}

A \textbf{situation} is an abstract description of either a state or action in the world.  A \textbf{curriculum} is a sequence of situations.   Situations are never accessed directly by the learner but rather only through the mediation of a derived perception (section \ref{sec:perceptual-representation}).

A situation consists of a set of \textbf{objects} which may have \textbf{properties}, a set of \textbf{relations} between these objects, and a set of \textbf{actions} which occur.  Object locations are not encoded via coordinates but only in terms of \textbf{regions} which are defined with respect to other objects; the way these regions are defined is largely drawn from \citeA{landau1993and}.  Objects, object properties, and relations are drawn from a user-defined \textbf{ontology}; this ontology is for the convenience of the curriculum designer only and is \emph{not} accessible to the language learner.  Certain aspects of a situation may be marked as \textbf{salient}, meaning they will be remarked upon in linguistic utterances derived from this situation.

More detail on the representation of situations (appendix  \ref{app:situation-representation-complex}) and ontologies (appendix \ref{app:ontologies}) may be found in the appendices.

\subsection{Situation Templates}
\label{sec:situation-templates}

\begin{figure*}[tbp]
    \centering
    \includegraphics{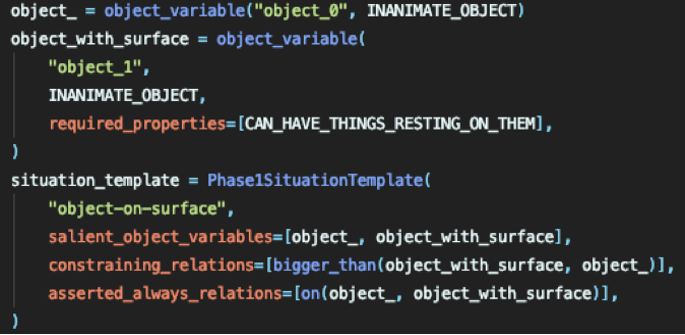}
    \caption{A simple situation template which will generate many concrete situations where an object is sitting on some flat surface.}
    \label{fig:basic-situation-template}
\end{figure*}

Defining every situation in a realistic curriculum consisting of thousands or millions of instances would be impractical. Instead, we allow researchers to use a simple Python-based domain-specific language to specify \textbf{situation templates} (Figure \ref{fig:basic-situation-template}).  For example, a situation template like \texttt{x[+ANIMATE, +HUMAN]} $\wedge$ \texttt{y[+ INANIMATE]} $\wedge$ \texttt{MANIPULATION\_ACTION(x, y)} $\wedge$ \texttt{RELATIVE\_SIZE(SMALLER, y, x)} could be instantiated by \texttt{PICK\_UP(PERSON, CUP)} or \texttt{ROLL(DOG, BALL)}.
  
  Situation templates have a similar structure to situations, but where situations contain objects, properties, relations, and actions situation templates can instead contain template variables which range over elements of the user-defined hierarchical ontology.  Restrictions can be placed on how these variables may be filled. A \textbf{situation generator} then reads these templates and instantiates the templates using either exhaustive enumeration or configurable sampling strategies.

Because situation templates are defined using Python code, the creation of templates can itself be automated. This can be useful, for example, to easily test all possible combinations of multiple experimental parameters.

 \subsection{Linguistic Representation}
\label{sec:linguistic-representation}
\label{sec:language-generator}

Any sequence of discrete symbols can in principle work as a linguistic representation for ADAM (phonemes, morphemes, etc.).  In the default implementation, we use English orthographic words\footnote{The plural marker is split off as its own token; learning morphonology could be done in an extended version of the default ADAM implementation} and Yale romanization for Chinese.

Linguistic representations are not directly specified as part of a curriculum but instead are generated from situation representations by a \textbf{language generator}.  This results in a key strength of ADAM: experiments can be performed in multiple languages from the same curriculum simply by swapping the language generator component.  The initial language generator implementations for English and Chinese are described in Appendix \ref{app:language-generator-implementations}.

\subsection{Perceptual Representation}
\label{sec:perception-generator}
\label{sec:perceptual-representation}

The language learner is not permitted to observe situations directly. Instead, a \textbf{perception generator} module first translates them to \textbf{perceptual representations}. The default implementation uses a perceptual representation designed to mimic those of a roughly 18-month-old child.  

This default representation includes object shape and structure representations derived from \cite{marr1982vision} and \cite{biederman1987recognition}; region and path information derived from \cite{landau1993and}; a small set of object properties, such as \texttt{liquid} and \texttt{self-moving}; the ability of the language learner to recognize particular familiar individuals; information on what objects are gazed at by the speaker \cite{conboy2008event}; very coarse grained relative size and possession relationships; which objects are the speaker and addressee; and proto-roles such as \texttt{stationary}, \texttt{causes-change}, and \texttt{volitionally-involved} derived from \cite{dowty1991thematic}.

The perception generator component could be swapped for a different implementation to run experiments with a different perceptual representation.  It is also possible to parameterize the current implementation to allow for the addition, exclusion, or unreliable perception of aspects of the representation.

\begin{table*}[ht]
\begin{center} 
\vskip 0.12in
\begin{tabular}{ll} 
Ontology Type & Ontology Subtypes \\
\hline
\hline
\textbf{thing}  & inanimate-object, head, hand, person, animal, \\
& body-part  \\
\hline
\textbf{person}  & me, you, Mom, Dad, baby  \\
\hline
\textbf{animal}  & dog, cat, bear, bird, chicken, cow  \\
\hline
\textbf{relation}  & in-region, spatial-relation, partOf, size-relation, \\
& axis-relation, has  \\
\hline
\textbf{action}  & walk, run, consume, put, push, shove, go, come, take,  \\
& grab, give, spin, sit, fall, throw, pass, move, jump, \\
& roll, fly  \\
\hline
\textbf{property}  & can-fill-template-slot, substance, is-addressee, \\
& is-speaker, is-human, perceivable-property, \\
& can-manipulate-objects, edible, rollable, \\
& can-have-things-on-them, is-body-part, \\
& person-can-have, transfer-of-possession, \\
& can-jump, can-fly, has-space-under, \\
& can-be-sat-on, fast, slow, side, left, right, \\
& hard-force, soft-force, aboutSameSizeAsLearner  \\
\hline
\textbf{color} & red, blue, green, black, white, light-brown, \\
& dark-brown, transparent  \\
\hline
\textbf{inanimate-object}  & substance, ground, food, table, bed, ball, \\
& paper, book, house, car, cup, box, chair, \\
& truck, door, hat, (furniture) leg, chairback, \\
& chairseat, tabletop, mattress, headboard, wall, \\
& roof, tire, truckcab, trailer, flatbed  \\
\hline
\end{tabular} 
\caption{The sample ontology included with ADAM. These ontology nodes describe scenes, which are used to generate plausible perception graphs with geons and other perceptual features, and then provided as input to the learner} 
\label{tab:sample-ontology} 
\end{center} 
\end{table*}

\subsection{Language Learner}
\label{sec:learning}

In ADAM, a \textbf{learner} is any object which can
\begin{itemize}
\item observe pairs of perceptual and linguistic representation and update its internal state in response.
\item observe a perceptual representation and produce a linguistic representation in response.
\end{itemize}

We intend ADAM as a ``sandbox'' where many different learning algorithms can be implemented.  By default, the system includes four learning algorithms: subset learning, Pursuit \cite{stevens2017pursuit},  Cross-Situational Learning \cite{cross-situational}, and Propose but Verify \cite{trueswell2013propose}.  These are described in detail in Appendix \ref{app:sub-learners}, along with a collection of representations and data structures for learning that can help researchers implement their own algorithms more easily. While the default learning algorithms are relatively simple and heuristic, arbitrary learning algorithms (e.g. deep neural networks, etc.) could be substituted.

\subsection{Evaluation}
\label{sec:evaluator}

The evaluation module allows the user to monitor learning progress (during training) and the effectiveness of the learned representations (at runtime). Details on this module are provided in Appendix \ref{app:experimentation}.

\section{Evaluation on a Sample Curriculum}
The ADAM system includes a simple sample curriculum.  Below, we describe the content of this curriculum and provide examples of what ADAM learns from it.

\subsection{Sample Curriculum Content}
The ontology of the sample curriculum is shown in Table \ref{tab:sample-ontology}. It consists of the following situations with:
\begin{itemize}
    \item single objects
    \item single objects sitting on the ground
    \item objects described by color
    \item objects possessed by the speaker
    \item objects possessed by a non-speaker
    \item objects resting upon other objects
    \item objects placed horizontally beside other objects
    \item objects positioned vertically relative to other objects
    \item objects inside other objects
    \item objects positioned in front of or behind other objects
    \item examples of all the verbs in the ontology
    \item examples of objects defined by their function (e.g. \textit{chair})
    \item multiple instances of the same type of object (plurals)
    \item objects with part-whole relationships
\end{itemize}

Some samples of what ADAM learns from the curriculum are given in figures \ref{fig:my} through \ref{fig:drink}. The representation used in these figures is described in Appendix \ref{app:perception-graphs}.


\section{Conclusions and Future Work}
\label{sec:conclusion}
In this report, we presented ADAM, a software system based in Python that can be used to simulate language acquisition experiments. Various system capabilities outlined in this report, including the modular abilities to craft learning curricula as desired (section \ref{sec:situation-representation-simple}) and to build different language learning algorithms (section \ref{sec:learning}) make ADAM suitable for systematic studies of language and how language acquisition models are impacted by the learning data, the perceptual complexity of the learning environment, and the cognitive resources available to the learner.

There are many future directions to improve the ADAM system. On the language processing side, the linguistic realization learned for objects, properties, relations, and actions could be enriched to have syntactic structure, rather than surface templates (Appendix \ref{app:surface-templates}). Also, the concept to language generation can be improved to take a situation and generate multiple potential language descriptions of the scene. Language processing can be done in real time using audio inputs for generating Speech-To-Text sequences. Similarly, perceptual processing can be further improved by processing real-world inputs from 2D or 3D visuals. This would be an undertaking to replace the perception generation module (Appendix \ref{app:perceptual-representation}) with a different processing system. 

Experiments with complex scenes is sometimes inconvenient due to ADAM's core graph matching algorithms being slow on some perception graphs.  To scale up experimentation, we need to shift the graph matching code from pure Python to a more efficient language and to explore newer graph matching algorithms than VF2.

\section{Acknowledgements}

Approved for public release; distribution is unlimited. This material is based upon work supported by the Defense Advanced Research Projects Agency (DARPA) under Agreement No. HR00111990060. The  views  and  conclusions  contained herein are those of the authors and should not be interpreted as necessarily representing the official policies or endorsements, either expressed or implied, of DARPA or the U.S. Government.

\bibliographystyle{apacite}

\setlength{\bibleftmargin}{.125in}
\setlength{\bibindent}{-\bibleftmargin}

\bibliography{adam}

\begin{figure*}
    \centering
    \includegraphics[scale=0.5]{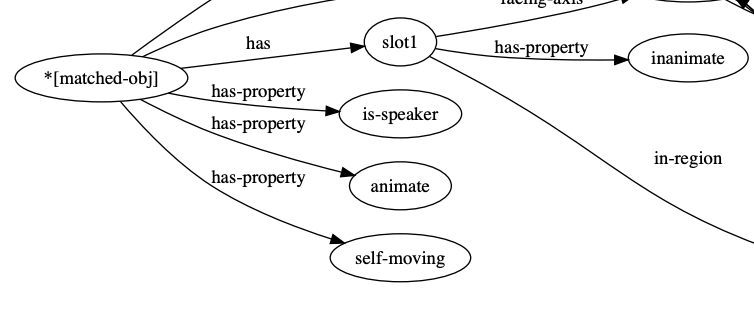}
    \caption{ADAM learns that \emph{my} indicates possession by the speaker.}
    \label{fig:my}
\end{figure*}

\begin{figure*}
    \centering
    \includegraphics[width=\textwidth]{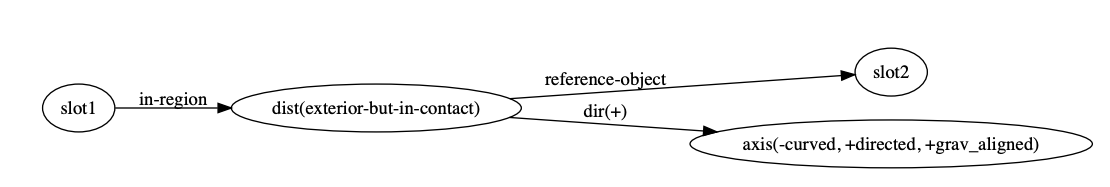}
    \caption{ADAM learns that for one object to be \emph{on} another, it needs to not only be above it, but also in contact with it.}
    \label{fig:on}
\end{figure*}

\begin{figure*}
    \centering
    \includegraphics[width=\textwidth]{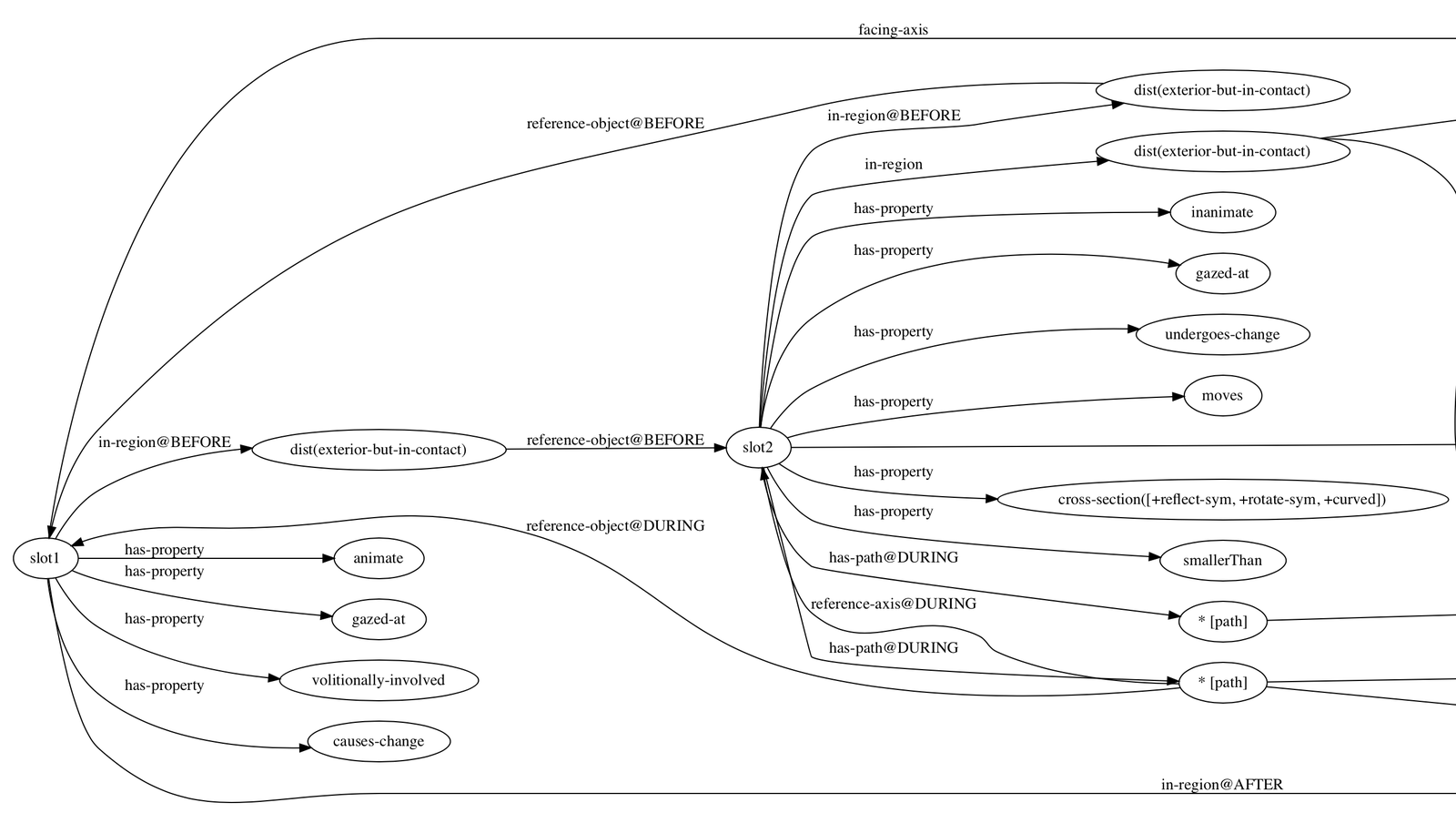}
    \caption{ADAM learns that you can only \emph{roll} things with circular cross-sections (see the \texttt{cross-section} property on \texttt{slot2})}
    \label{fig:roll}
\end{figure*}

\begin{figure*}
    \centering
    \includegraphics[scale=0.37]{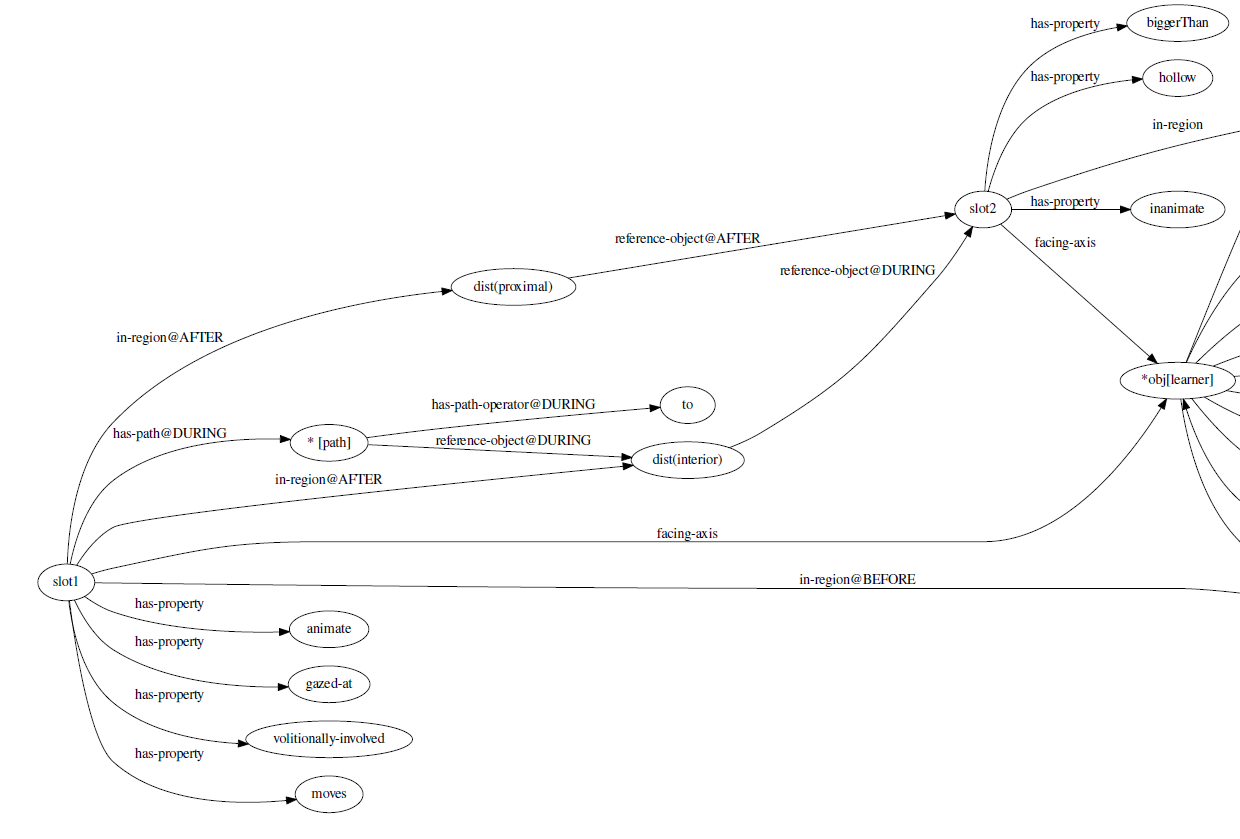}
    \caption{ADAM learns that \emph{goes in} involves the movement of an object into the interior of a hollow object.}
    \label{fig:goes-in}
\end{figure*}

\begin{figure*}
    \centering
    \includegraphics[scale=0.6]{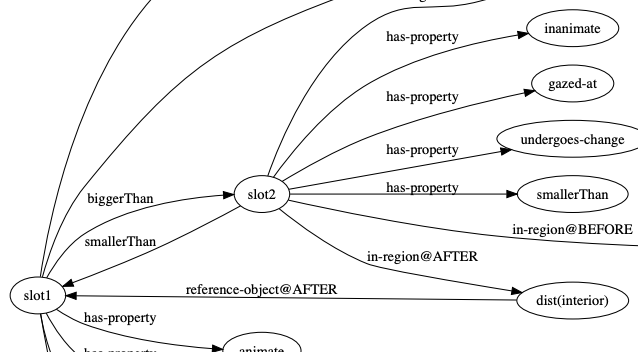}
    \caption{ADAM learns that if someone is eating something, the eater is animate, while the eatee is inanimate and smaller than the eater, and the eatee becomes interior to the eater}
    \label{fig:eat}
\end{figure*}

\begin{figure*}
    \centering
    \includegraphics[width=\textwidth]{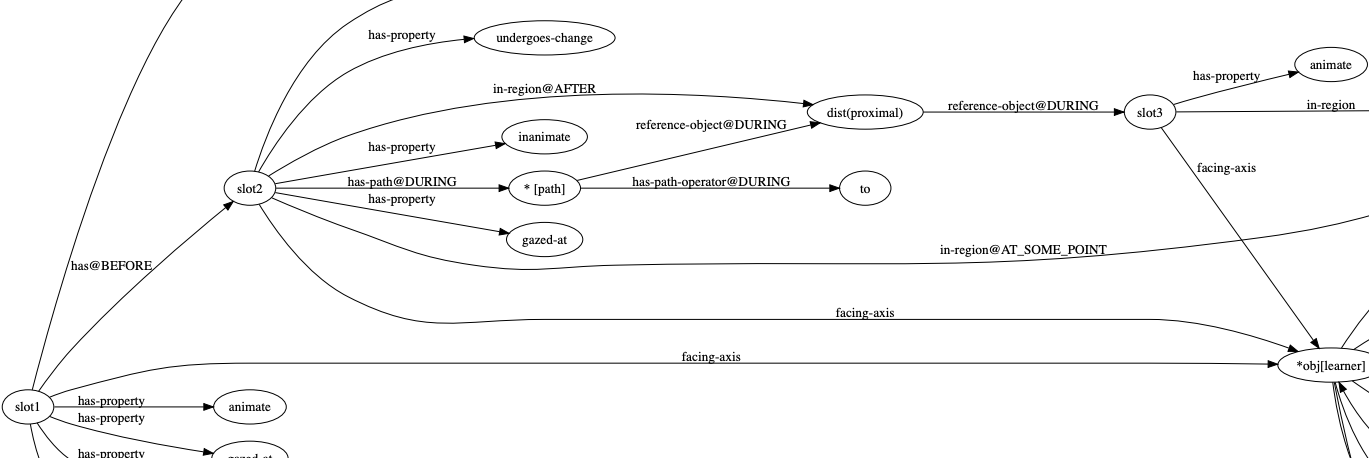}
    \caption{ADAM learns that the thrower loses possession of the object, which traverses a path above the ground towards someone else.  It does not learn that the catcher necessarily gains possession of the ball (e.g. \emph{The quarterback threw the ball to the open man, but the receiver dropped it})}
    \label{fig:throw-to}
\end{figure*}

\begin{figure*}
    \centering
    \includegraphics[width=\textwidth]{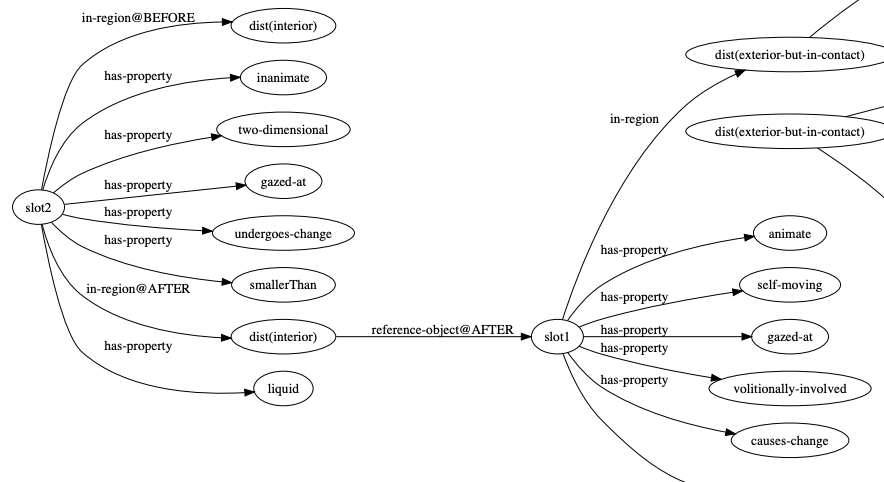}
    \caption{ADAM learns you can only drink liquids.}
    \label{fig:drink}
\end{figure*}

\clearpage

\begin{appendices}

\section{Ontologies}
\label{app:ontologies}

An \textbf{ontology} is a collection of object, attribute, relation, and action types used for two purposes:
\begin{itemize}
    \item to define the objects, attributes, relations, and actions in a situation and to provide the necessary information for transforming them into the perceptual representation
    \item to define how variables in situation templates may be instantiated.  For example, the arguments of a certain action in a template may be restricted to animate objects.
\end{itemize}

An ontology is structured as a collection of \textbf{ontology nodes} with parent-child relationships.  Every node may be associated with a set of \textbf{ontology node properties} which are inherited by all its child nodes.  There are six special nodes which must be present in any ontology.  \textsc{thing}, \textsc{property}, \textsc{relation}, and \textsc{action} are self-explanatory.  \textsc{perceivable} is a meta-property used to mark which other properties can be perceived by the learner.  \textsc{can\_fill\_template\_slot} indicates those nodes which can potentially be arguments of relations and actions.  

An ontology node derived from \textsc{object} may be associated with an \textbf{object structural scheme} which provides a hierarchical representation of the internal structure of a type object and guides how that object is perceived as a collection of sub-objects and geons.  A structural schema represents the general pattern of the structure of an object, rather than the structure of any particular object.  For example a person's body is made up of a head, torso, left arm, right arm, left leg, and right leg. These sub-objects have various relations to one another (e.g. the head is above and supported by the torso).

Any ontology node derived from \textsc{action} may be associated with an \textbf{action description} which guides how that action is perceived in terms of changes to objects and their relations.  An action description consists of:
\begin{itemize}
    \item A mapping of semantic roles to \textbf{action description variables}. All the relations and conditions specified below will be in terms of these variables, which will be bound to concrete objects when an action is instantiated in a particular situation.
    \item a collection of pre-conditions, which are relations which hold only in the frame before the action happens
    \item a collection of post-conditions, which are relations which hold only in the frame after the action happens
    \item a collection of enduring conditions, which are relations which hold before and after the action happens.
    \item a description of what happens during the action, consisting of
    \begin{itemize}
        \item a mapping from action description variables to paths traversed by those objects during the action
        \item a set of relations which hold continuously during the action
        \item a set of relations which hold at some point, but not necessarily continuously, during the action.
    \end{itemize}
    \item properties required on objects involved in the action (e.g. the \textsc{agent} must be \textsc{animate}
    \item \textbf{auxiliary variables}, which do not occupy semantic roles but are still referred to by conditions, paths, etc.  An example would be the container for the \textsc{liquid} which is an argument of a \textsc{drink} action.
\end{itemize}

\section{Situation Representation}
\label{app:situation-representation-complex}

A situation consists of 
\begin{itemize}
    \item a set of objects, some subset of which may be marked as \textbf{salient}. These are the ones which will be foregrounded in any linguistic descriptions of the situation. For example, the ground is always present but rarely salient.  Every object in a situation possesses a type from the ontology and three axes; these axes can be used to construct regions to describe the spatial relationships of objects to one another in an abstract way.
    \item a set of relations between objects in the situation which hold before and/or after any actions occur.  It is not necessary to state every relationship which holds in a situation. Rather this should contain the salient relationships which should be expressed in the linguistic description.
    It is also not necessary to state relations which are implied by actions occurring in the situation.
    \item which objects in the situation are gazed at by the speaker.
    \item a set of \textbf{actions} which occur in the situation.  An action is an action description from the ontology together with a mapping from the semantic roles and auxiliary variables of the action to the objects in the situation.
    \item \textbf{syntax hints}, which can constrain how the language generator expresses the situation.  A curriculum designer can, for example, use these to force actions in a situations to always be described in the passive voice, if possible.
\end{itemize}

\section{Regions}
\label{app:regions}

Regions (which we derive almost entirely from \cite{landau1993and}'s account of the semantics of prepositions) are used to describe the locations of objects relative to one another in an abstract way.  They may be used in action descriptions, situations, perceptions, and situation templates.

A \textbf{region} consists of:
\begin{itemize}
    \item a reference object
    \item a \textbf{distance}, one of \textsc{interior}, \textsc{exterior\_but\_in\_contact}, \textsc{proximal}, and \textsc{distal}.
    \item a \textbf{direction}, which is an object axis together with a positive or negative polarity.
\end{itemize}

Regions can also be used to construct paths which describe how objects move.  A \textbf{path} (whose representation we again derived almost entirely from \cite{landau1993and}):
\begin{itemize}
    \item source and destination reference object or regions
    \item an optional \textbf{path operator}, which must be one of \textsc{via}, \textsc{to}, \textsc{toward}, \textsc{from}, 
    or \textsc{away\_from}
    \item an optional reference axis
    \item whether the orientation of an object changes as it moves along the path
\end{itemize}

\section{Situation Templates}
\label{app:situation-templates}

A \textbf{situation template} has the same structure as a template, except all objects, properties, relations, and actions may be replaced by variables. The instantiations of these variable is controlled by constraints, the most common being restrictions on which types from the ontology can be used.

Situation templates may be instantiated into concrete situations using either a sampling or exhaustive enumeration strategy.

\section{Initial Language Generator Implementations}
\label{app:language-generator-implementations}

ADAM by default provides language generators for English and Chinese.  These implementations both work by translating the portions of situations marked as salient into dependency trees.  The learner has no access to these dependency trees but only the resulting token sequence.

\section{Default Perceptual Representation}
\label{app:perceptual-representation}

The world is represented to the learner as either one or two \textbf{perception frames} (one for static situations; two for those containing actions, encoding the state of the world before and after the action).
A perception frame consists of a collection of \textbf{perception objects} which have \textbf{properties} and have \textbf{relations} with one another. 
Perception objects
\begin{itemize}
    \item may have sub-objects with which they have a \textsc{sub-part} relation.
    \item have three \textbf{axes}, one of which is marked as \textsc{primary} (typically the longest). 
    These axes may have relations between each other (for example, to indicate that an object is longer along one axis than another). 
    Each axis has boolean flags to indicate whether it is
    \begin{itemize}
        \item curved
        \item directed 
        \item aligned with gravity
    \end{itemize}
    \item a \textbf{geon} (with a few exceptions, such as objects which appear two-dimensional). A geon has
    \begin{itemize}
        \item three axes.  Axes can have coarse-grained size relationships with each other (e.g. for a pencil, its extent along one axis is much greater than the other two, which are roughly the same).
        \item a \textbf{generating axis}, which usually corresponds to its longest axis.
        \item a \textbf{cross-section} data structure which specifies for the cross-section of the object along its primary axis
        \begin{itemize}
            \item whether it is curved
            \item whether it has reflective symmetry
            \item whether it has rotational symmetry
            \item its size, which is one of
            \begin{itemize}
                \item \textsc{constant} if the cross-section is of roughly constant size along the generating axis
                \item \textsc{small-to-large} if the cross-section gets bigger along the positive direction of the generating axis
                \item \textsc{large-to-small} if the cross-section gets smaller along the positive direction of the generating axis
                \item \textsc{small-to-large-to-small} if the cross section starts small, gets large in the middle, and then shrinks (along the generating axis).
            \end{itemize}
        \end{itemize}
    \end{itemize}
    \item may have \textbf{properties}. The following properties can be perceived:
    \begin{itemize}
        \item  \textsc{self-moving}
        \item \textsc{animate}
        \item \textsc{inanimate}
        \item \textsc{two-dimensional}
        \item \textsc{liquid}
        \item \textsc{hollow}
        \item \textsc{recognized-particular} (\textsc{is-dad}, \textsc{is-mom}, \textsc{is-baby}, \textsc{is-learner}, \textsc{is-ground})
        \item \textsc{gazed-at}
        \item Proto-roles \cite{dowty1991thematic}
        \begin{itemize}
            \item \textsc{volitionally-involved}
                    \item \textsc{sentient-or-perceives}
            \item \textsc{causes-change}
            \item \textsc{moves}
            \item \textsc{undergoes-change}
                    \item \textsc{incremental-theme} (roughly speaking, this is the object involved in an action whose part-whole state corresponds to the progress of the action itself, e.g. \emph{sandwich} in \emph{eat a sandwich})
            \item \textsc{stationary}
                    \item \textsc{causally-affected}

        \end{itemize}
        \item colors, expressed as an red-green-blue triples
    \end{itemize}
    \item may have a \textbf{path} (see \ref{app:ontologies}).
\end{itemize}

The following relations are perceivable
\begin{itemize}
    \item \textsc{part-of}
    \item \textsc{bigger-than}, \textsc{smaller-than}, \textsc{much-bigger-than}, \textsc{much-smaller-than}
    \item \textsc{possession}
    \item \textsc{in-region} (which takes a region as an argument)
\end{itemize}

Two-frame perceptions of actions can also contain descriptions of what happens during the action itself, consisting of:
\begin{itemize}
        \item a mapping from objects to paths (\ref{app:regions}) traversed by those objects during the action
        \item a set of relations which hold continuously during the action
        \item a set of relations which hold at some point, but not necessarily continuously, during the action.
\end{itemize}

Perceptions are generated from situations by combining the situation representations with information from the ontology such as ontology node properties, object structural schemata, and action descriptions.

\section{Learning Representations}
\label{app:learning-representations}

ADAM provides representations and algorithms used by all the sample learning algorithms implemented by ADAM (Appendix \ref{app:learning-algorithms}).  Although user-implemented algorithms are not required to use them, they do make the process of writing a learning algorithm much simpler.

\subsection{Perception Graphs}
\label{app:perception-graphs}
\begin{figure*}[tbp]
    \centering
    \includegraphics[width=\textwidth]{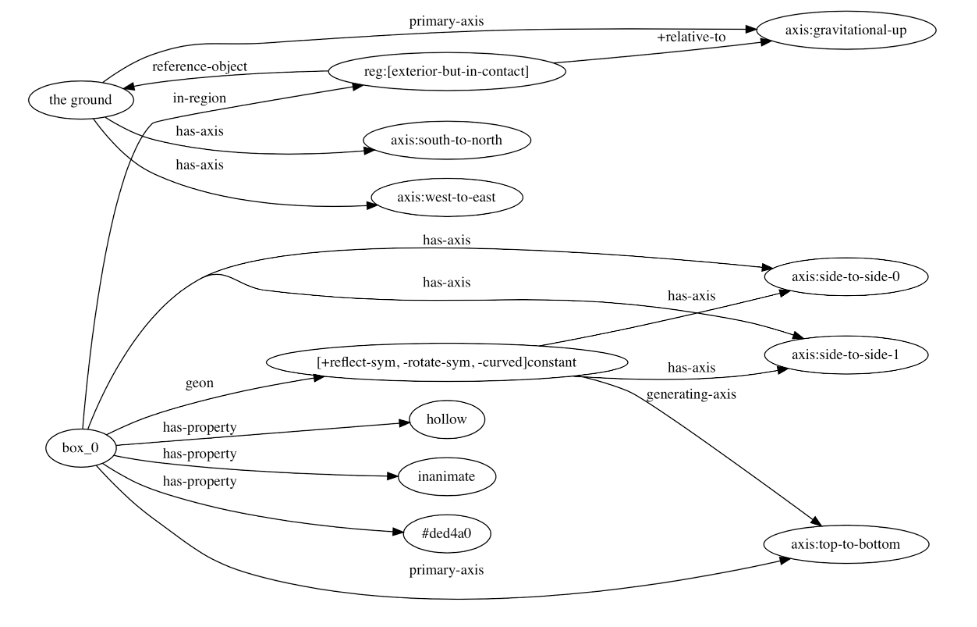}
    \caption{A simple perception graph for the learner’s perception of a box sitting on the ground.}
    \label{fig:perception-graph}
\end{figure*}

For our learning algorithms, it is convenient to convert our learner’s perceptual representation into a directed acyclic graph (a \textbf{perception graph}; Figure \ref{fig:perception-graph}). A node in a perception graph can represent:
\begin{itemize}
    \item an object perception (Appendix \ref{app:perceptual-representation}) 
    \item geons (Appendix \ref{app:perceptual-representation})
    \item an object property (Appendix \ref{app:perceptual-representation})
    \item a spatial region (Appendix \ref{app:regions})
    \item an axis (Appendix \ref{app:regions})
    \item a path (Appendix \ref{app:regions})
\end{itemize}
Note that some nodes, such as object perceptions and axes, have labels on them for the convenience of developers when debugging (\texttt{box\_0}, \texttt{axis:side-to-side-0}). These are not accessible to the learner.

Edges between nodes represent relations from the inventory in Section \ref{sec:perceptual-representation} (e.g. \textsc{partOf}).  
Geons and paths are represented as multi-node sub-graphs, with each aspect of their structure described in Appendix \ref{app:perceptual-representation} getting its own node, joined by special edge types.
In dynamic situations, every edge in a perception graph is labelled with a set of \textbf{temporal scopes} (either \textsc{before}, \textsc{after}, or both) indicating at which frame(s) during the action the relationship encoded by the edges holds true.

\subsubsection{Perception Graph Patterns}
\label{app:perception-graph-patterns}

\begin{figure*}[tbp]
    \centering
    \includegraphics[scale=1]{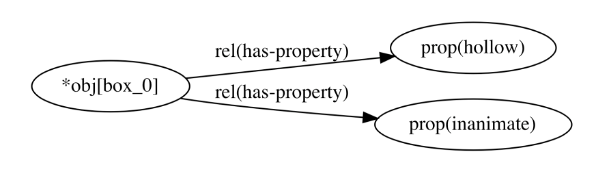}
    \caption{An example of a perception graph pattern which would match any perception graph that has an object perception node with properties \textsc{hollow} and \textsc{inanimate}, such as that of a box or a cup.}
    \label{fig:perception-graph-pattern}
\end{figure*}

ADAM's default learner represents the ``meaning'' of an object, relation, or action by \textbf{perception graph patterns} which can be matched against the perception graphs for new situations.  Like perception graphs, these are directed acyclic graphs, but their nodes and edge are boolean functions (\textbf{predicates}) which accept or reject matches against nodes and edges of perception graphs, respectively (Figure \ref{fig:perception-graph-pattern}).

The predicates can be complex; for example, we can require an edge have both a certain type and a certain temporal scope. Patterns can also contain special \textbf{slot variable} nodes; these are `wildcards' corresponding to the arguments of the property, relation, or action the pattern represents.

\subsection{Perception Graph Algorithms}
\subsubsection{Perception Graph Pattern Matching}
Matching a perceptual graph pattern against a perception graph becomes an instance of a restricted form of the sub-graph isomorphism problem. Although this problem is NP-complete in general, in practice it runs efficiently over the sort of graphs encountered in ADAM. 
ADAM uses the implementation of the VF2 sub-graph isomorphism algorithm \cite{cordella2001improved, cordella2004sub} from the \texttt{networkx} library \cite{SciPyProceedings_11} with a few of custom optimizations.

ADAM also supports matching perception graphs patterns against one another (e.g. to determine if one pattern is a generalization rather than another pattern). This can also be accomplished by a slight variation on the same algorithm by imposing a partial order on the node and edge predicates according to the set of nodes and edges they can possibly match.

\subsubsection{Perception Graph Pattern Generalization}
ADAM learning algorithms encounter the need to generalize patterns in one of two ways:
\begin{itemize}
    \item Given a pattern and perception graph the pattern fails to match, determine if there is some generalization of the pattern which would successfully match the graph.
    \item Given two patterns, compute their intersection (the largest sub-pattern which is implied by both input patterns). 
\end{itemize}

Both of these can be approximated using this greedy match generalization algorithm:
\begin{itemize}
    \item Attempt a regular match. If successful, stop.
    \item If the match failed, use information gathered during the match process to determine which pattern node blocked the expansion of the largest alignment found between pattern nodes and graph nodes. `Relax' this node either by deleting it or (in a few cases) performing a node-specific relaxation (e.g. if the node was a color predicate, we might expand the range of colors it is willing to match).  
    \item When certain nodes are deleted, the deletion is propagated to other graph nodes:
    \begin{itemize}
        \item if an object is deleted, so are its sub-objects
        \item if during relaxation a portion of the graph becomes (weakly-)disconnected and contains no slot variables, remove it.
    \end{itemize}
\end{itemize}

\subsection{Surface Templates and Perception Graph Templates}
\label{app:surface-templates}
A \textbf{surface template} is a sequence of tokens or argument slots, such as \emph{ARG1 gives ARG2 to ARG3} or \emph{green ARG1}. 
The argument slots are referred to as \textbf{surface template variables}. 
A surface template can be instantiated to produce a token sequence by providing a binding of surface template variables to tokens.
Alternate representations with richer syntactic structure can easily be introduced.

A perception graph template is a perception graph pattern together with an alignment of its slot variable nodes to surface template variables. 
the linguistic realization of learned objects, properties, relations, and actions (including objects, which are degenerate argumentless templates). 

\subsubsection{Concepts}

ADAM's default implementation uses an extremely simplified semantic representation consisting of a set of \textbf{concepts}, which are of four types:
\begin{itemize}
    \item object concepts, which have no argument slots
    \item attribute concepts, which take one other concept as an argument
    \item relation concepts, which take two other concepts as (numbered) arguments
    \item action concepts, which take a variable number of other concepts as (numbered) arguments
\end{itemize}

\subsubsection{Alignment Data Structures}
\label{app:alignments}

A \textbf{language-concept alignment} (LCA) is a mapping between concepts and their corresponding token sequences.  A \textbf{perception-concept alignment} (PCA) is a mapping between concepts and sub-graphs in a perception graph.  A \textbf{language-perception-concept alignment} (LPCA) is pair of an LCA and a PCA over the same set of concepts.

During training, the sample integrated learner implementation builds up an LPCA. At inference time, it builds up a PCA and then translates it into language.

\section{Sample Learners}
\label{app:learning-algorithms}
\subsection{Top-level Integrated learner}
In the sample implementation, learning is coordinated at the top-level by an \textbf{integrated learner} which contains sub-learners for objects, attributes, relations, and actions, each of which can be customized to use different learning algorithms.
The integrated learner first initializes an empty LPCA (\ref{app:alignments}) to represent its understanding of the relationship between the linguistic utterance and the perception graph.
It then applies each sub-learner in the order objects, then attributes, then relations, then actions.  
Each sub-learner observes the linguistic utterances, the perception graph, and the LPCA.  After updating its internal state, each learner then has the opportunity to update the LPCA before it is passed down the pipeline.  
For example, if the object learner is confident it has learned the concept \textsc{dog} and it recognizes one in the perception graph, it can align \textsc{dog} to the token \emph{dog} in the utterance and to the corresponding portion of the perception graph.

At inference time, the same process is applied, except that a PCA is used instead of an LPCA.  After all sub-learners have finished, the PCA is discarded and the set of concepts is translated to language using the algorithm below, which will be illustrated using the following concepts: \{ \textsc{dad}, \textsc{ball}, \textsc{throws}(\textsc{dad}, \textsc{ball}), \textsc{red}(\textsc{ball}), \textsc{shiny} (\textsc{ball} ) \}
\begin{enumerate}
    \item each object concept is translated to a set of token sequences as follows:
    \begin{itemize}
        \item first, get the (argument-less) token sequence learned for the object concept by the object sub-learner (e.g.\textsc{ball} to \emph{ball}). If there is none, the object concept is not translated.  In the initial implementation, ADAM does not try to learn English determiners and they are added by special logic.
        \item second, get the set of surface templates associated with the attributes\footnote{This should also be done for relations, but this has not yet been implemented.} of this object by the attribute sub-learner (if any).  For example, \{\emph{red} X, \emph{shiny} X\}.
        \item return a set consisting of the result of recursively instantiating all subsets (of size $<=3$) of the relation surface templates (using the object surface template to bind the argument slot of the first attribute surface template).  For example, \{\emph{ball}, \emph{shiny ball}, \emph{red ball}, \emph{shiny red ball}\}
    \end{itemize}
    \item each relation concept mapped to a surface template by the relation sub-learner is translated to a set of token sequences by instantiating the surface template with its two arguments bound to each pair from the cross-product of all instantiations of the argument object concepts (from step 1).
    \item each action concept mapped to a surface template by the action sub-learner is translated to a set of token sequences by instantiating its the surface template with its arguments bound to each tuple from the cross-product of all instantiations of the argument object concepts (from step 1). For example, \emph{Dad throws the ball}, \emph{Dad throws the red ball}, \emph{Dad throws the shiny ball}, \emph{Dad throws the shiny red ball}.

\end{enumerate}

The scope of language covered by this process is limited and enhancing this is a direction for future work.

\subsection{Sub-Learners}
\label{app:sub-learners}

All of our sample sub-learners share a common structure.  First, all learners except the object learner preprocess the perception graph to replace all sub-graphs aligned to object concepts with single nodes annotated with those concepts.  Next,  candidate surface templates are generated from the language-concept alignment:
\begin{itemize}
    \item for objects, any unaligned token.
    \item for attributes, any unaligned token before or after a token aligned to an object concept forms a candidate surface template (e.g. if \emph{red ball on the table} is the utterance and \emph{ball} is the only aligned token, then \emph{red ARG1} and \emph{X on} would be candidates).
    \item for relations, any span of unaligned tokens for length at most 2 between two aligned tokens aligned to object concepts is a candidate.
    \item for verbs, ADAM uses a complex set of patterns to generate surface templates based which account for different possible word order patterns in a language.
\end{itemize}
Finally, for each candidate surface template one or more perception hypotheses is created for each candidate surface template:
\begin{itemize}
    \item for objects, any object perception subgraph not already aligned is a candidate.
    \item for attributes, only single property perception nodes attached to the arguments are candidates.
    \item for relations, the single initial hypothesis is the sub-graph consisting of all shortest paths between the arguments together with any nodes one hop off this path
    \item for actions, the initial hypothesis is the entire graph.
\end{itemize}
If there was not an existing hypothesis for this surface template from previously observed situations, these new hypotheses are stored for it.  Otherwise, this new hypothesis set is used to update the existing hypothesis set in a way which depends on the particular learning algorithm; this can cause the hypothesis set to become empty, in which case the learner concludes the corresponding surface pattern candidate should be discarded (e.g. the learner may eventually conclude that \emph{red} is not an action).

The following sample learning algorithms are provided in ADAM:
\begin{itemize}
    \item \textbf{Subset}.  This learns the `meaning' of a word to be those aspects of the perception graph which are present every time it is used (inspired by \cite{webster1989automatic}).
    It is provided as a simple baseline, but it will fail in the presence of noise.  
    ADAM's sample implementation uses it for learning actions.
    \item \textbf{Pursuit}, which implements the word learning model of \cite{stevens2017pursuit} into a variant we call \textsc{pursuit-na}. (see appendix \ref{app:pursuit-na})
    It learns well under noisy conditions, but struggles with complex hypothesis spaces.
    ADAM's sample implementation uses it for object, attribute, and relation learning.
    \item ADAM includes a less-tested implementation of Cross-Situational Learning \cite{cross-situational}, a global learning model that uses distributional statistics across learning instances to make hypotheses for word meanings.
    \item Propose but Verify \cite{trueswell2013propose}, a local learning model that proposes and retains a word meaning hypothesis after a single word-learning instance, and abandons the hypothesis if the subsequent learning instances do not confirm it is also provided. Similarly to the Cross-Situation Learning model, this is also less tested.
\end{itemize}

\subsection{Pursuit Non-Atomic}
\label{app:pursuit-na}
The \textbf{Pursuit} algorithm was designed to learn an alignment between words and atomic meanings, as is typically the case in word learning experiments. However, the grounded world as perceived by the learner in ADAM is not made up of simple atomic objects. Rather objects form sub-graphs of a rich, complex perceptual structure. This requires that the meaning representations assigned to words must have a similarly rich and complex structure. We use ADAM perception patterns (appendix \ref{app:perception-graph-patterns}) for this representation.  

Adapting \textbf{Pursuit} to use perception patterns as meaning representations (resulting in a variant we call \textsc{Pursuit-na} for `non-atomic') requires several changes to the base algorithm.  First, \textsc{Pursuit-na} needs to select a new hypothesis from the intractably large space of possible perception sub-graphs both in (1) when a new word is encountered (to choose the initial hypothesis) and in (3) when rewarding a new random hypothesis because the leading hypothesis is disconfirmed. Fortunately, ADAM's perceptual representation assumes the boundaries between objects (but not their identities) are known at a pre-linguistic stage \cite{Spelke2007}.  Our implementation of \textsc{pursuit-na} therefore uses these object boundaries to define  the hypothesis space whenever it needs to make random choices.

Second, the perceptual representation of an object will contain both elements essential to the meaning of the corresponding word and elements which are accidental.  For example, the shape of a ball is crucial to the meaning of the word \emph{ball}, but its color is not; however, from the perspective of a child, the color of \emph{juice} may be essential to how they use the word. Without a way to generalize our hypotheses and distinguish the essential from the accidental, \textsc{pursuit-na} will be unable to learn because its leading hypotheses will consistently fail to match in situations outside those where they were initially perceived.  To address this, \textsc{pursuit-na} modifies \textsc{Pursuit} in two ways:
\begin{itemize}
    \item  when attempting to confirm whether or not the current leading hypothesis matches a new situation in (2), \textsc{pursuit-na} exploits ADAM's ability to find partial matches of patterns. If a perfect match is not found, \textsc{pursuit-na} still disconfirms and penalizes the leading hypotheses. However, instead of rewarding a random hypothesis in (3), the sub-graph of the leading hypothesis which did successfully match the scene's perception graph is rewarded, so long as the number of nodes in the matched sub-pattern is of sufficient size relative to the original pattern (this size ratio is the \emph{partial match parameter}).  For example, if the learner believes being red is part of the meaning of \emph{ball} and then observes the word \emph{ball} being used in a situation with a green ball, it will reward the (possibly new) hypothesis which contains the shape information but omits the color information. 
    \item  Whenever any hypothesis is rewarded (in (2) or (3)), \textsc{pursuit-na} also rewards any hypotheses which are generalizations of that hypothesis, where one pattern is a generalization of another if its perception pattern graph representation is a sub-graph of the other's. This helps prevent the learner from learning very slowly by continually jumping between overly specific hypotheses, with more general hypotheses taking a very long time to become the leading hypothesis, which is necessary for their scores to increase. 
\end{itemize}

\subsection{Specialized Learners}
ADAM provides two specialized learners which are modules that augment the normal sub-learners to provide additional learning capability. These specialized learner's operate on ADAM's perceptual representation (see Appendix \ref{app:perceptual-representation}) either before or after the other sub-learners discussed in Appendix \ref{app:sub-learners}. These learners target linguistic structures which depend on actions, relations, attributes, or objects to construct a representation of.

\begin{itemize}
    \item \textbf{Plural Learner} that learns templates for plurals (e.g many cats, two balls). Although the plural learner module learns linguistic templates very similarly to the attribute learner module that learn attributes such as colors, it runs only when the perceived scene contains multiple objects that match the same pattern. The module preprocesses the scene to mark these objects perceptually with nodes indicating counts to be learned as count attributes.
    \item \textbf{Functional Object Learner} that learns meanings of objects based on their function (e.g. \textit{a person sits on a \textunderscore{chair}}. An independent model. unlike the standard object learner, this module tracks the interactions between a specific known object concept and how the object is used in an action concept. Allowing the learner to make an assumption to what an unknown object functioning in the same roll of an action might be. For example, if a chair in the action \emph{“A man sits on a chair”} is replaced by a chair with only three legs, or a stool which has no back the learner could still describe the action as \emph{“A man sits on a chair”} even though the specific object is not immediately recognized correctly.
\end{itemize}

\section{Experimentation and Evaluation Framework}
\label{app:experimentation}

ADAM provides a convenient framework for setting up and executing experiments.
The user specifies:
\begin{itemize}
    \item the language learner to use. This may optionally be initialized from a saved learner state (e.g. to run experiments with actions with a learner which has already learned the objects appearing in the actions).
    \item one or more training stages. A training (or test) \textbf{stage} is a sequence of $(situation, linguistic utterances, perceptual representation)$ tuples. Typically these are not provided directly; instead the user provides a collection of situation templates, a template instantiation strategy (e.g. random sampling), a language generator, and a perception generator.
    \item zero or more test warm-up stages. These are observed by the learner without evaluation before evaluating on the test set.
    \item one or more test stages.  
    \item a collection of \textbf{experiment observers}. These observers can hook into different points in the experiment process (before and after the observation of each training example, after training is complete, after each test instance).
\end{itemize}

Experiment observers are provided to:
\begin{itemize}
    \item track how often the top-scoring linguistic description produce by the learner on an instance matches the `gold' utterance associated with the instance.  
    \item track how often the `gold' utterance associated with an instance appears anywhere in the learner's candidate description list.
    This is typically more useful to track since there are numerous valid and reasonable linguistic utterances in any given situation.
    \item write a detailed report of the learning process to a human-readable HTML file.
    This report contains the situation, linguistic utterance, and perceptual representation of every training and test instance, together with the linguistic utterances produced by the learner.
\end{itemize}

\end{appendices}

\end{document}